\pdfoutput=1
\documentclass[11pt,a4paper]{article}

\usepackage[hyperref]{emnlp-ijcnlp-2019}
\usepackage{times}
\usepackage{latexsym}
\usepackage{array,multirow}
\usepackage{graphicx}
\usepackage{amsmath}
\usepackage{color}

\DeclareMathOperator*{\argmin}{argmin}

\usepackage{url}

\aclfinalcopy 

\title{“Transforming” Delete, Retrieve, Generate Approach for Controlled Text Style Transfer}

\author{
  Akhilesh Sudhakar, Bhargav Upadhyay, Arjun Maheswaran\\
  Agara\\
  \tt{\{akhilesh, bhargav, arjun\}@agaralabs.com}
}

\date{}

\begin{document}
\maketitle
\begin{abstract}
Text style transfer is the task of transferring the style of text having certain stylistic attributes, while preserving non-stylistic or content information. In this work we introduce the Generative Style Transformer (GST) - a new approach to rewriting sentences to a target style in the absence of parallel style corpora. GST leverages the power of both, large unsupervised pre-trained language models as well as the Transformer.  GST is a part of a larger `Delete Retrieve Generate' framework, in which we also propose a novel method of deleting style attributes from the source sentence by exploiting the inner workings of the Transformer. Our models outperform state-of-art systems across 5 datasets on sentiment, gender and political slant transfer. We also propose the use of the GLEU metric as an automatic metric of evaluation of style transfer, which we found to compare better with human ratings than the predominantly used BLEU score.
\end{abstract}

\section{Introduction}
Text style transfer is an important Natural Language Generation (NLG) task, and has wide-ranging applications from adapting conversational style in dialogue agents \cite{zhou2017mechanism}, obfuscating personal attributes (such as gender) to prevent privacy intrusion \cite{reddy-knight-2016-obfuscating}, altering texts to be more formal or informal \cite{rao-tetreault-2018-dear}, to generating poetry \cite{yang-etal-2018-stylistic}. The main challenge faced in building style transfer systems is the lack of parallel corpora between sentences of a particular style and sentences of another, such that sentences in a pair differ only in style and not content (non-stylistic part of the sentence). This has given rise to methods that circumvent the need for such parallel corpora.  

Previous approaches using non-parallel corpora, that employ learned latent representations to disentangle style and content from sentences, are typically adversarially trained \citep{pmlr-v70-hu17e,shen2017style, fu2018style}. However these models a) are hard to train and take long to converge, b) need to be re-trained from scratch to change the trade-off between content retention and style transfer c) suffer from sparsity of latent disentangled representations, d) produce sentences of bad quality (according to human ratings) and e) do not offer fine-grained control over target style attributes.

\citet{li-etal-2018-delete} find that style attributes are more often than not, localized to a small subset of words of a sentence. Building on this inductive bias, they model style transfer in a ``Delete Retrieve Generate" framework (hereby referred to as DRG) which aims to 1) delete only the set of attribute words from a sentence to give the content, 2) retrieve attribute words from the target style corpus, and 3) use a neural editor (an encoder-decoder LSTM) to generate the final sentence from the content and retrieved attributes.

While DRG as a framework leads to output sentences that are better in quality than previous approaches, their individual Delete and Generate methods are susceptible to: a) removing core content words which would preserve crucial context, b) failing to remove source style attributes that should be replaced with target style attributes, c) the LSTM-based encoder-decoder model not being robust to errors made by the Delete and Retrieve models, d) generating sentences that are not fluent, by abruptly forcing retrieved attributes into the source sentence and e) failing on longer input sentences.

In this work, we propose a novel approach to rewrite sentences into a target style, that leverages the power of both a) transfer learning by using an unsupervised language model trained on a large corpus of unlabeled text, as well as b) the Transformer \citep{NIPS2017_7181}. We refer to our Transformer as the Generative Style Transformer \textbf{(GST)}.
We use the DRG framework proposed by \citet{li-etal-2018-delete} but we overcome the shortcomings of their a) Delete mechanism, by using the attention weights of another Transformer that we refer to as the Delete Transformer \textbf{(DT)}, and b) Generate mechanism by using \textbf{GST}, which does away with the need for (and consequent shortfalls of) a sequence-to-sequence encoder-decoder architecture using LSTMs.

We outperform the current state-of-art systems on transfer of a) sentiment\footnote{We use style in a broad socio-linguistic sense that encompasses sentiment too, for the purpose of this work}, b) gender and c) political slant. Our approach is advantageous in that it is simple, controllable and exploits the important inductive bias described, while at the same time it leverages the power of Transformers in novel ways. 

All code, data and results for this work can be found in our Github repository \footnote{https://github.com/agaralabs/transformer-drg-style-transfer}.

\section{Our Approach}
Given a dataset $ D =  \{(x_1, s_1), . . . ,(x_m, s_m)\}$ where $x_i$ is a sentence and $s_i \in S$ is a specific style, our goal is to learn a conditional distribution $ P(y | x, s^{tgt}) $ such that $ Style(y) = s^{tgt} $, where style is determined by an oracle that can accurately determine the style of a given sentence. For instance, for the sentiment transfer task, $S = \{$'Positive', 'Negative'$\}$. Using the DRG framework, we model our task in 3 steps: \newline
\textbf{(1)} A \textbf{Delete} model which learns $P(c,a|x)$ such that $c$ and $a$ are non-stylistic and stylistic components of $x$ respectively, $ Style(c) \notin S $ (i.e., $c$ does not have any particular style) and $x$ can be completely reconstructed from $c$ and $a$, \textbf{(2)} A \textbf{Retrieve} model which retrieves a set of (optional) target attributes $a^{tgt}$ from $D_{s^{tgt}}$, the corpus of sentences of target style, and \textbf{(3)} A \textbf{Generate} model in two flavors: a) one which learns to generate a sentence in the target distribution $P(y|c, s^{tgt})$ and b) another which learns to generate a sentence in the target distribution $P(y|c, a^{tgt})$, both such that $Style(y)=s^{tgt}$.
We now elaborate on each of these components individually.

\subsection{Delete}
For an input sentence ``The restaurant was \textit{big} and \textit{spacious}", in the case of a style transfer task from positive to negative sentiment, the Delete model should be capable of deleting the style attributes \textit{big} and \textit{spacious}. 

Our approach to attribute deletion is based on `input reduction' \citep{feng-etal-2018-pathologies}, based on the observation that certain words and phrases significantly contribute to the style of a sentence. For a sentence $x$ of style $s_j$ having a set of attributes $a$, a style classifier will be confused about its style if the attributes in $a$ are removed from $x$. We describe a mechanism to assign an \textit{importance} score to each token in $x$, which is reflective of its contribution to style. These scores allows us to distinguish style attributes from content.

\subsubsection{Delete Transformer}
To build intuition, any attention-based style classifier defines a probability distribution over style labels: 
\begin{equation} \label{eq:att_wts_lstm_eq} 
p(s|x) = g(v, \alpha) 
\end{equation}
where $v$ is a tensor such that $v[i]$ is an encoding of $x[i]$, and $\alpha$ is a tensor of attention weights such that $\alpha[i]$ is the weight attributed to $v[i]$ by the classifier in deciding probabilities for each $s_j$. The $\alpha$ scores can be treated as \textit{importance} scores and be used to identify attribute words, (which typically tend to have higher scores). 
Motivated by the recent successes of the Transformer \citep{NIPS2017_7181} and more specifically, BERT \citep{devlin2018bert}, on a number of text classification tasks (including achieving state-of-art results on sentiment classification), we use a BERT-based transformer as our style classifier and refer to it as Delete Transformer \textbf{(DT)}. However, since \textbf{DT} has multiple attention heads and multiple blocks (layers), extracting a single set of attention weights $\alpha$ is a non-trivial task. This is further complicated by the fact that every layer and head encodes different aspects of semantic and linguistic structure \citep{vig2019visualizing}. We then use a novel method to extract a specific attention head and layer combination that encodes style information and that can be directly used as \textit{importance} scores. \newline\newline
\textbf{Attribute extraction:}
We use the same input representation as Figure 3(b) in \citet{devlin2018bert} wherein a `[CLS]' token is added before the sentence tokens. Since the softmax classification layer is used over the attention stack of the `[CLS]' token in BERT classifiers, the attention weights of other input tokens that correspond to `[CLS]' are of special interest in identifying significant sentence tokens. First, we iterate over each pair $<h, l>$ (head-layer pair) and extract the attention scores for every token $w$ of $x$ as follows:
\begin{equation} \label{salience}
\alpha_{h,l}(w) = softmax_{w \in x}(Q_{h,l,[CLS]}K_{h,l,w}^{T})
\end{equation}
where `Q' and `K' carry the same original connotations of query and key vectors as used by \citet{NIPS2017_7181}, in the Transformer as:
\begin{equation} \label{transformer_att}
Att(Q, K, V) = softmax(\frac{QK^{T}}{ \sqrt{d_k}})V
\end{equation}
We then remove the top $\gamma|x|$ tokens from $x$, based on \textit{importance} scores calculated as in Eq. \ref{salience}. Keeping in line with \citet{feng-etal-2018-pathologies}, we call this removal a `reduction', and denote the resulting reduced sentence as $x'_{h,l}$. $\gamma$ is a parameter we tune to each dataset which allows us to control the proportion of words in a sentence to be deleted, and $|x|$ denotes the number of tokens in $x$. We calculate a score $z(x'_{h,l})$:
\begin{equation} \label{z_score} 
z(x'_{h,l}) = \frac{p(s|x'_{h,l})+\lambda}{\sum_{s'}p(s'|x'_{h,l})+ \lambda}
\end{equation}
where $\lambda$ is a smoothing parameter, $s$ is the style label assigned maximum probability by the softmax distribution over all styles in the label set $S$, and $s' = S - \{s\}$.
The final pair $<h_{s}, l_{s}>$ out of combinations of all heads H and layers L, is obtained by averaging the score in Eq. \ref{z_score} over a validation set of `reduced' sentences $D'_{val}$ as follows:
\begin{equation} \label{z_score_avg} 
(h_{s},l_{s}) = \argmin_{h \in H, l \in L} \frac{\sum_{x'_{h,l} \in D_{val}'}z(x'_{h,l})}{|D_{val}'|}
\end{equation}
A `reduction' of any input sentence $x$ based on $<h_{s}, l_{s}>$ gives us $x'_{h_{s},l_{s}}$ which we refer to as the content $c$. The removed tokens are the attributes $a$.\newline\newline
\textbf{Evaluation of Extracted Attributes:} We evaluate our Delete method using human evaluation on Amazon Mechanical Turk\footnote{https://www.mturk.com/}, on which annotators were asked to choose if all the style-related attributes are extracted correctly by our Delete mechanism, and if any non-style attributes are wrongly deleted. We used 200 random sentences from our test set for sentiment transfer, for this evaluation. Our method deleted all style attributes on 89\% of examples, and wrongly deleted non-style attributes only 12\% of the time. In comparison, the Delete mechanism proposed by \citet{li-etal-2018-delete} deleted all style attributes only 67\% of the time, and wrongly deleted non-style attributes over 29\% of the time. \newline

\subsection{Retrieve}
We retrieve a sentence from the target style corpus of sentences according to:
\begin{equation} \label{retrieve_eq}
    x^{tgt} = argmin_{x' \in D_{s^{tgt}}} d(c_{x}, c_{x'})
\end{equation}
where $d$ is a distance metric, such that contents which are closer according to $d$ will have compatible attributes as they occur in similar contexts. We experiment with multiple retrieval mechanisms, using cosine similarity over different sentence representations: a) TF-IDF weighted, b) Averaged-GloVe over all tokens of a sentence and c) Universal Sentence Encoder \citep{cer2018universal}. We obtain best retrieval results using TF-IDF vector similarity.

\begin{figure*}[htp]
\centering
\includegraphics[width=12cm]{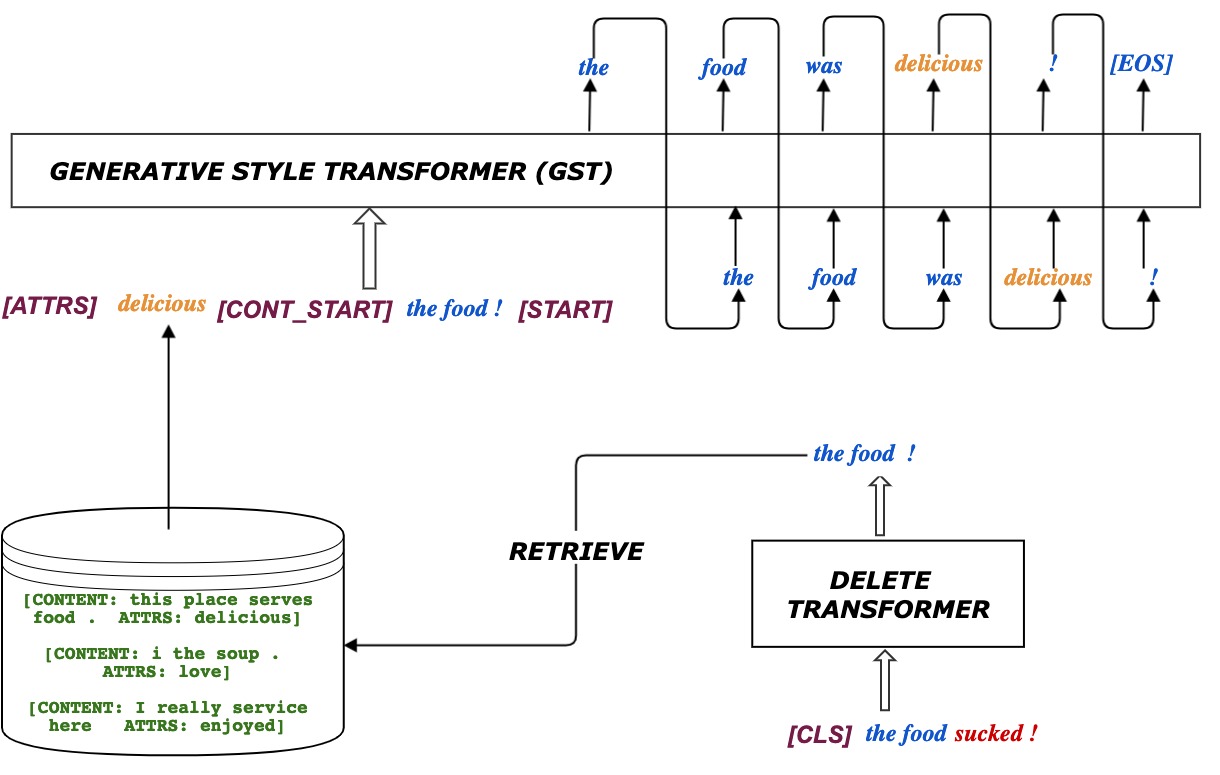}
\caption{Our architecture, with an example from the Yelp dataset for the task of sentiment transfer}
\label{end2end}
\end{figure*}

\subsection{Generate}

Our approach to generate sentences of the target style leverages both the power of transfer learning by using an unsupervised language model trained on a large corpus of unlabeled text, as well as the Transformer model. The model we use is a multi-layer `decoder-only' Transformer which is based on the Generative Pre-trained Transformer (GPT) of \citet{radford2018improving}. This is our Generative Style Transformer \textbf{(GST)}. \textbf{GST} has masked attention heads that enable it to look only at the tokens to its left, and not to those to its right. \textbf{GST} derives inspiration from the fact that recently, many large generatively pre-trained Transformer models have shown state-of-art performance upon being finetuned on a number of downstream tasks. It is trained to learn a representation of content words and (retrieved) attribute words presented to it, and generate fluent sentences in the domain of the target style while attending to both content and attribute words appropriately. 

\subsubsection{Variants of GST (B-GST and G-GST)}
Taking cues from \citet{li-etal-2018-delete}, we train GST in two flavors: the Blind Generative Style Transformer \textbf{(B-GST)} and the Guided Generative Style Transformer \textbf{(G-GST)}. For a sentence $x$ of a source style $s^{src}$ with content $c$ and retrieved (target style) attributes $a^{tgt}$, the two variations are learnt as follows.\newline 
\textbf{B-GST:} The inputs to this model are only $c$, and $s^{tgt}$. The output \textbf{$y$} of the model is the generated sentence in the target style. In this setting, the model is free to generate the output sentence, given the content and the target style, and is \textbf{blind} to specific desired target attributes. \textbf{B-GST} can be useful in cases when the target corpus does not have sentences that are similar to the source corpus, which causes the Retrieve model to retrieve incompatible target attributes.\newline\newline
\textbf{G-GST:} The inputs to this model are $c$, and $a^{tgt}$, and the output \textbf{$y$} of the model is the generated sentence in the target style. In this setting, the model is \textbf{guided} towards generating a target sentence with desired attributes. \textbf{G-GST} is useful for two reasons. Firstly, in cases when the target corpus has similar sentences to the source corpus, it reduces sparsity by giving the model information of target attributes. Secondly, and more importantly, it allows fine-grained control of output generation by manually specifying target attributes that we desire during inference time, without even using the Retrieve component. This controllability is an important feature of G-GST that most other latent-representation based style transfer approaches do not offer. 

\subsubsection{Input Representation and Output Decoding}
Taking inspiration from \citet{devlin2018bert}, we add special tokens to indicate target style, and to indicate the demarcation between content and attributes. For \textbf{B-GST} the input at timestep $t$ of target sentence prediction consists of special tokens to denote: a) target style $s^{tgt}$, b) the start of content $c$, d) the start of output, followed by all target tokens up till and including timestep $t-1$. \textbf{G-GST} has a similar input representation, except that a special token to indicate start of retrieved attributes is added, and the retrieved attributes are provided before the content. The target style $s^{tgt}$ is not provided. Our end-to-end architecture for \textbf{G-GST} is depicted in Figure \ref{end2end}, including input representation. \textbf{B-GST} is similar in nature, except that it does not use a retrieve component. At timestep $t$, both \textbf{GST}s predict the $t^{th}$ output token,  by generating a probability distribution over words in the vocabulary according to: a) $p(y_t | c, y_{1}, y_{2}, .. y_{t-1})$ for \textbf{B-GST}, and b) $p(y_t | c,  a^{tgt}, y_{1}, y_{2}, .. y_{t-1})$ for \textbf{G-GST}. This is done by using a softmax layer over the topmost Transformer block corresponding to $y_{t-1}$. During training time, we use the `teacher forcing' or `guided approach' \citep{Bengio:2015:SSS:2969239.2969370, Williams89alearning} over decoded tokens. During test time, we beam search using softmax probabilities with a look-left window of 1 and a beam width of 5. The output beam (out of the top 5 final beams) that obtains the highest target-style match score using the Delete Transformer described earlier, is chosen as the output sentence.

\subsubsection{Training}
Since we do not have a parallel corpus, both GSTs are trained to minimize the reconstruction loss. Specifically, for a sentence $x$, the model learns to reconstruct $y = x$ given $c_{x}$, its own attributes $a_{x}$ (only for \textbf{G-GST}) and its own style $s^{src}$ (only for \textbf{B-GST}). More formally \textbf{B-GST} learns to maximize the following objective:
\begin{equation}
L(\theta) = \sum_{(x,s^{src})\in D} \log p(x | c_{x}, s^{src}; \theta)\newline
\end{equation}
However, training \textbf{G-GST} using the reconstruction loss in this manner results in the model learning to trivially combine $c_x$ and $a_x$ to generate $x$ back. In reality we want it to be capable of \textit{adapting} target attributes into the context of the source content, in a non-trivial manner to produce a fluent sentence in the target style. To this end, we noise the inputs of the \textbf{G-GST} model during training time, by choosing random attributes for 10\% of the examples (5\% from the source style and 5\% from the target style), to replace $a_{x}$. Denoting the chosen attribute for an example (either noisy or its own) to be  $a'_{x}$, \textbf{G-GST} learns to maximize the following objective:
\begin{equation}
L(\theta) = \sum_{(x,s^{src})\in D} \log p(x | c_{x}, a'_{x}; \theta)\newline
\end{equation}

\subsubsection{Model Details and Pre-training}
We use the PyTorch implementation of the pre-trained Transformer by HuggingFace\footnote{https://github.com/huggingface/pytorch-pretrained-BERT}, which uses the pre-trained OpenAI GPT model\footnote{https://github.com/openai/finetune-transformer-lm}. This model is pre-trained by \citet{radford2018improving} on the BookCorpus dataset\footnote{https://www.smashwords.com/} of over 7000 books (around 800M words). \textbf{GST} has a sequence length of 512, 12 blocks (or layers), and 12 attention-heads in each block. All internal states (keys, queries, values, word embeddings, positional embeddings) are 768-dimensional. Input text is tokenized using Byte-Pair Encoding (BPE).


\section{Experiments} 

\subsection{Datasets}

We use 5 different datasets for our experiments. These datasets have been used in previous works on style transfer.\newline\newline
We use the YELP, AMAZON and CAPTIONS datasets as used by \citet{li-etal-2018-delete}, and we retain the same train-dev-test split that they use. Further, they also provide human gold standard references for the test sets of all 3 of the above.
We use the POLITICAL \citep{rtgender} and GENDER \cite{reddy-knight-2016-obfuscating} datasets as used by \citet{style_transfer_acl18}. We have retained the same train-dev-test split that they use. Table \ref{dataset-stats} shows statistics of these datasets. Following are brief descriptions of the datasets:

\noindent
\textbf{YELP:} Reviews of businesses on Yelp, with each review labelled as having either positive or negative sentiment.\newline
\textbf{AMAZON:} Product reviews on Amazon, with each review labelled as having either positive or negative sentiment.\newline
\textbf{CAPTIONS:} Image captions, with each caption labeled as either factual, romantic, or humorous. The task is to
convert factual sentences into romantic and humorous ones. 
\newline
\textbf{POLITICAL:} Top-level comments on Facebook posts from members of the United States Senate and House who have public Facebook pages, with each comment labelled as having been posted by either a Republican or a Democrat politician.\newline
\textbf{GENDER:} Reviews of food businesses on Yelp, with each review labelled as either of the two genders (male or female) corresponding to markers of sex.\newline

\begin{table}
\begin{center}
\small
\begin{tabular}{|c|c|c|c|c|}
\hline \bf Dataset & \bf Style & \bf Train & \bf Dev & \bf Test \\ \hline
YELP & Positive & 270K & 2000 & 500 \\
& Negative & 180K & 2000 & 500 \\
\hline
CAPTIONS & Romantic & 6000 & 300 & 0 \\
& Humorous & 6000 & 300 & 0 \\
& Factual & 0 & 0 & 300 \\
\hline
AMAZON & Positive & 277K & 985 & 500 \\
& Negative & 279K & 1015 & 500 \\
\hline
POLITICAL & Democrat & 270K & 2000 & 28K \\
& Republican & 270K & 2000 & 28K \\
\hline
GENDER & Male & 1.34M & 2250 & 267K \\
& Female & 1.34M & 2250 & 267K \\
\hline
\end{tabular}
\end{center}
\caption{\label{dataset-stats} Dataset statistics}
\end{table}


\begin{table*}
\begin{center}
\small
\begin{tabular}{|c|c|c|c|c|c|c|c|c|c|c|c|c|}
\hline
& \multicolumn{4}{|c|}{YELP} & \multicolumn{4}{|c|}{AMAZON} & \multicolumn{4}{|c|}{CAPTIONS} \\
\hline
\hline
Model & Cont. & Flu. & Sty. & All & Cont. & Flu. & Sty. & All & Cont. & Flu. & Sty. & All \\
\hline
D\&R & 18 & 14.5 & 22 & 13 & 40 & 35.5 & 45 & 39.5 & 20.5 & 24 & \textbf{47.75} & 30.25\\
\textbf{B-GST} & \textbf{66} & \textbf{64} & \textbf{60} &\textbf{69} & \textbf{48} & \textbf{50.5} & \textbf{45.5} & \textbf{49} & \textbf{65.5} & \textbf{56.75} & 34 & \textbf{52.75} \\
\hline
None & 16 & 21.5 & 18 & 18 & 12 & 14 & 9.5 & 11.5 & 14 & 19.25 & 18.5 & 17\\
\hline
\end{tabular}
\end{center}
\caption{Human evaluation results - each cell indicates the percentage of sentences preferred down a column (Cont. = Content preservation ; Flu. = Fluency ; Sty. = Target Style Match ; All = Overall)}
\label{YAI-human-DR-BGST}
\end{table*}


\begin{table}
\begin{center}
\small
\begin{tabular}{|c|c|c|c|c|c|c|}
\hline
& \multicolumn{2}{|c|}{POLITICAL} & \multicolumn{2}{|c|}{GENDER} \\
\hline
\hline
Model & Cont. & Flu. & Cont. & Flu.\\
\hline
BT & 22 & 29 & 18 & 23\\
\textbf{B-GST} & \textbf{69} & \textbf{61} & \textbf{70} &\textbf{67} \\
\hline
None & 9 & 10 & 12 & 10\\
\hline
\end{tabular}
\end{center}
\caption{Human evaluation results - each cell indicates the percentage of sentences preferred down a column (Cont. = Content preservation ; Flu. = Fluency)}
\label{PG-human-DR-BGST}
\end{table}

\subsection{Comparison to Previous Works}
On the Yelp, Amazon, and Captions dataset, we compare with 3 previous adversarially trained models: StyleEmbedding \textbf{(SE)} \citep{fu2018style}, MultiDecoder \textbf{(MD)} \citep{fu2018style}, CrossAligned \textbf{(CA)} \citep{shen2017style}) and the 2 best models - DeleteOnly \textbf{(D)} and DeleteAndRetrieve \textbf{(D\&R)} of \citet{li-etal-2018-delete} trained using the DRG framework. A brief description of the first 3, and a detailed description of the last 2 models can be found in \citet{li-etal-2018-delete}, so we omit elaborating on them here. At the time of writing this paper, these models are the top performing models on Yelp, Amazon and Captions, with the \textbf{D\&R} model of \citet{li-etal-2018-delete} showing state-of-art performance. Output sentences of each of these 5 models on fixed test sets, also annotated with human reference gold standards \textbf{(H)} on all 3 datasets are provided by \citet{li-etal-2018-delete}. We use the same for our comparison and evaluation. On the Political and Gender datasets, we compare our models against that of \citet{style_transfer_acl18}, which is the state-of-art on these 2 datasets at the time of writing this paper. Their trained models for both these datasets are made publicly available. They use back-translation \textbf{(BT)} using an LSTM as a mechanism to learn latent representations of source sentences, and then employ adversarial generation techniques to make the output match a desired style \citep{style_transfer_acl18}. 

\section{Evaluation of Results}

The widely agreed upon goals for a style transfer system are 1) Content preservation of the non-stylistic parts of the source sentence, 2) Style transfer strength of the stylistic attributes to the target style and 3) Fluency and correct grammar of the generated target sentence \citep{mir2019evaluating}. To this end, we use both human and automatic evaluation to measure model performance.
\subsection{Human Evaluation}
\textbf{YELP, AMAZON and CAPTIONS:}
\citet{li-etal-2018-delete} report state-of-art results
which we corroborate through manual and automatic metrics. We then proceed to obtain human evaluations on these models along with ours through Amazon Mechanical Turk\footnote{https://www.mturk.com/}.
Specifically, we ask annotators to rate each pair of generated sentences given the source sentence, on content preservation, style transfer strength, fluency, and overall success. For each parameter, they are asked to choose which of the generated sentences is better, or neither of the two if they are unable to decide. Table \ref{YAI-human-DR-BGST} presents results on our best scoring model \textbf{B-GST} with the previous best scoring model \textbf{D\&R} as a percentage of times one was preferred over the other. \newline
\textbf{POLITICAL and GENDER:} 
On these 2 datasets, \cite{style_transfer_acl18} report state-of-art results using their model \textbf{BT}, which we similarly corroborate. A comparison of our best model \textbf{B-GST}, with their results using \textbf{BT} is presented in Table \ref{PG-human-DR-BGST} as a percentage of times one was preferred over the other. Since judging target style strength on these two tasks are hard for MTurkers, they only rate these datasets for content and fluency.


\begin{table*}
\begin{center}
\small
\begin{tabular}{|c|c|c|c|c|c|c|c|c|c|c|c|c|}
\hline
& \multicolumn{4}{|c|}{YELP} & \multicolumn{4}{|c|}{AMAZON} & \multicolumn{4}{|c|}{CAPTIONS} \\
\hline
\hline
Model & GL & BL\textsubscript{S} & PL & AC & GL &  BL\textsubscript{S} & PL & AC & GL &  BL\textsubscript{S} & PL & AC \\
\hline
\hline
SRC & 7.6 & 100.0 & 24.0 & 2.6 & 19.3 & 100.0 & 32.9 & 20.4 & 11.3 & 100.0 & 34.4 & 50.0 \\
\hline
\hline
CA & 4.4 & 48.0 & 72.8 & 72.7 & 0.0 & 15.2 & \textbf{30.1} & \textbf{83.1} & 1.6 & 24.1 & \textbf{10.1} & 50.8 \\
SE & 5.9 & 78.0 & 115.9 & 8.6 & 0.0 & 16.7 & 129.8 & 45.5 & 5.9 & 53.8 & 80.3 & 51.0 \\
MD & 5.0 & 57.3 & 205.6 & 46.8 & 0.0 & 16.5 & 122.5 & 71.8 & 4.5 & 48.7 & 40.5 & 51.3 \\
\hline
D & 6.4 & 56.7 & 75.8 & 85.0 & 0.0 & 16.2 & 55.0 & 50.6 & 7.8 & 59.1 & 52.5 & 57.5 \\
D\&R & 6.9 & 58.0 & 90.0 & \textbf{89.3} & 0.0 & 16.1 & 42.2 & 50.9 & 7.8 & 49.1 & 28.8 & \textbf{67.5} \\
\hline
\textbf{G-GST} & 3.8 & 70.6 & 64.4 & 78.3 & 13.4 & 71.0 & 171.0 & 57.6 & 1.1 & 13.1 & 45.0 & 52.3 \\
\textbf{B-GST} & \textbf{11.6} & \textbf{71.0} & \textbf{38.6} & 87.3 & \textbf{14.9} & \textbf{73.6} & 55.2 & 60.0 & \textbf{12.6} & \textbf{68.3} & 28.9 & 56.0 \\
\hline
\hline
H & 100.0 & 58.1 & 67.2 & 75.2 & 100.0 & 70.5 & 77.0 & 42.6 & 100.0 & 36.4 & 41.4 & 55.5 \\
\hline

\hline
\end{tabular}
\end{center}
\caption{Automatic evaluation results (GL = GLEU, BL\textsubscript{s} = BLEU ; PL = Perplexity ; AC = Target Style Accuracy ; SRC = Input Sentence ; B-GST and G-GST are our models ; H = Human Reference)}
\label{YAI-auto-DR-BGST}
\end{table*}

\begin{table}
\begin{center}
\small
\begin{tabular}{|c|c|c|c|c|c|c|}
\hline
& \multicolumn{3}{|c|}{POLITICAL} & \multicolumn{3}{|c|}{GENDER}\\
\hline
\hline
Model & BL\textsubscript{S} & PL & AC & BL\textsubscript{S} & PL & AC\\
\hline
\hline
SRC & 100.0 & 62.9 & 9 & 100 & 183.4 & 18.9 \\
\hline
\hline
BT & 40.2 & \textbf{61.9} & \textbf{88.0} & 46.0 & 196.2 & 52.9 \\
\hline
\textbf{G-GST} & 76.7 & 241.6 & 67.4 & 78.5 & 252.0 & 49.0 \\
\textbf{B-GST} & \textbf{79.2} & 104.4 & 71.2 & \textbf{82.5} & \textbf{189.2} & \textbf{57.9} \\
\hline
\end{tabular}
\end{center}
\caption{Automatic evaluation results (BL\textsubscript{s} = BLEU; PL = Perplexity; AC = Target Style Accuracy; SRC = Input Sentence; B-GST and G-GST are our models) }
\label{PG-auto-DR-BGST}
\end{table}

\subsection{Automatic Evaluation}
As has been done by previous works, we attempt to use automatic methods of evaluation to assess the performance of different models. To estimate target style strength, we use style classifiers that we train on the same training-dev-test split of Table \ref{dataset-stats}, using FastText\footnote{https://fasttext.cc/} \citep{joulin2017bag}. These classifiers achieve 98\%, 86\%, 80\%, 92\% and 82\% accuracies on the test sets of Yelp, Amazon, Captions, Political and Gender respectively. 
To measure content preservation, we calculate the BLEU score \citep{Papineni2001BleuAM} between the generated and source sentences. To measure fluency, we finetune a large pre-trained language model, OpenAI GPT-2 (note that this is different from GPT-1 on which our Generate model is based) on the target sentences using the same training-dev-test split of Table \ref{dataset-stats}. We use this language model to measure perplexity of generated sentences. The language models achieve perplexities of 24, 33, 34, 63 and 81 on the test sets of Yelp, Amazon, Captions, Political and Gender respectively. As we analyze in the next section, automatic metrics are inadequate at measuring the success of a good style transfer system.\newline
\textbf{GLEU:} As a step towards finding an automatic metric that compares with human judgements, we propose the use of the Generalized Language Evaluation Understanding Metric (GLEU) \citep{napoles-etal-2015-ground} - originally proposed as a grammatical error correction (GEC) metric. In the interest of space, we omit writing the elaborate equations and explanation for GLEU in this paper, but instead point the reader to Section 4 of \citet{napoles-etal-2015-ground} for the same. The formulation of GEC is quite similar to our formulation of style transfer in that style transfer involves making localized edits to the input sentence. Unlike BLEU, which takes only the target reference and the generated output into consideration, GLEU considers both of these as well as the source sentence too. It is a suitable metric for style transfer because it a) penalizes words of the source that were wrongly changed in the generated sentence,  b) rewards words that were successfully changed and c) rewards those that were successfully retained from the source sentence to match those in the reference sentence. We use the implementation of GLEU\footnote{https://github.com/cnap/gec-ranking} provided by \citet{napoles-etal-2015-ground}.

\newcommand{\specialcell}[2][c]{%
  \begin{tabular}[#1]{@{}c@{}}#2\end{tabular}}

Tables \ref{YAI-auto-DR-BGST} and \ref{PG-auto-DR-BGST} show a comparison of automatic metrics between our models and previous models described earlier.

\begin{table*}
\begin{center}
\small
\begin{tabular}{|c||c|c|}
\hline
\textbf{Example \#1} & \textbf{YELP ({\color{blue} Positive} to {\color{red} Negative})} & \textbf{YELP ({\color{red} Negative} to {\color{blue} Positive})}\\
\hline
\hline
SRC & \specialcell[t]{i {\color{blue} love} this place , the service is always {\color{blue} great} !} &   \specialcell[t]{the store is {\color{red} dumpy} looking \\ and management {\color{red} needs to change } .} \\
\hline
\hline
SE & \specialcell[t]{i {\color{blue} love} this place , the service is always {\color{blue} great} !} &   \specialcell[t]{the store is bought \\ the building does n't deal .} \\
\hline
D & \specialcell[t]{i paid \_num\_ minutes before the gifted , \\ not , a huge plus , n't .} &  \specialcell[t]{the store is {\color{red} dumpy} looking and management 
\\ is {\color{blue} fantastic} and {\color{red} needs to change} .}\\
\hline
D\&R & \specialcell[t]{i did {\color{red} not like} the homework of lasagna , \\ {\color{red} not like} it , .} & \specialcell[t]{the store is {\color{blue} clean} and {\color{blue} well} {\color{red} dumpy} looking
\\ and management {\color{red} needs to change .}} \\
\hline
\textbf{G-GST} & \specialcell[t]{\textbf{i used this place , the service is always {\color{red} awful} !}} & \specialcell[t]{\textbf{the store is looking and } \\
\textbf{management is {\color{blue} excellent} to .}}\\
\hline
\textbf{B-GST} & \specialcell[t]{\textbf{i {\color{red} hate} this place , the service is always {\color{red} terrible} !}} & \specialcell[t]{\textbf{the store is looking {\color{blue} great} and } \\ \textbf{management to {\color{blue} perfection} .}} \\
\hline
\\
\hline
\textbf{Example \#2}& \textbf{AMAZON ({\color{blue} Positive} to {\color{red} Negative})} & \textbf{AMAZON ({\color{red} Negative} to {\color{blue} Positive})}\\
\hline
\hline
SRC & \specialcell[t]{i finally made he purchase and am {\color{blue} glad} i did .}  & \specialcell[t]{i m just looking forward to the day \\ i get to {\color{red}replace} it .}\\
\hline
\hline
SE & i \specialcell[t]{finally made he purchase and am {\color{blue} glad} i did .}  & \specialcell[t]{i m just looking forward to the \\ {\color{blue}right away i get it} for it .}\\
\hline
D & i \specialcell[t]{finally made it and was {\color{blue} excited} \\ to purchase and am {\color{blue} glad} i did .} & \specialcell[t]{i m just looking forward to the day \\ i get to {\color{red}replace} my old one .}\\
\hline
D\&R & \specialcell[t]{i finally made i will try another \\ purchase and am {\color{blue} glad} i did .} & \specialcell[t]{looking forward to {\color{blue} using} it on turkey day !  .}\\
\hline
\textbf{G-GST} & \specialcell[t]{\textbf{i finally made he purchase and am} \\ \textbf{{\color{red} embarrassed} i {\color{red} smell pungent .}}} & \specialcell[t]{\textbf{i m looking to the same day} \\ \textbf{i get to {\color{blue}use} it .}}\\
\hline
\textbf{B-GST} & \specialcell[t]{\textbf{i finally made him purchase and am {\color{red} sorry} i did .}} & \specialcell[t]{\textbf{i m looking forward to {\color{blue} using}} \\ \textbf{the day i get to {\color{blue} use} it .}} \\
\hline
\\
\hline
\textbf{Example \#3}& \textbf{CAPTIONS (Factual to {\color{red} Romantic})} & \textbf{CAPTIONS (Factual to {\color{blue} Humorous})}\\
\hline
\hline
SRC &  \specialcell[t]{people gather around a life size chess game .} & three  \specialcell[t]{brown and black dogs are \\ splashing in the water .}\\
\hline
\hline
SE & \specialcell[t]{ people gather at a red wooden\\ advertisement of players {\color{red} enjoy} .} & \specialcell[t]{three small and brown dog \\are splashing in the water .} \\
\hline
D & \specialcell[t]{two young boys have working around a dream line \\and {\color{red}dream} of childhood.}
  & \specialcell[t]{three black and brown dogs are sitting in the water \\{\color{blue}to search of fish .}} \\
\hline
D\&R & \specialcell[t]{people gather around a carnival event ,\\ all determined to win the game .} & \specialcell[t]{two black and brown dogs are running \\in the water {\color{blue} like a fish.}}\\
\hline
\textbf{G-GST} &  \specialcell[t]{\textbf{people gather around a life size chess game} \\ \textbf{to celebrate {\color{red} life ' s happiness} .}} & \specialcell[t]{\textbf{three brown and black dogs are splashing in} \\\textbf{ the water {\color{blue} talking to each other} .}}\\
\hline
\textbf{B-GST} &  \specialcell[t]{\textbf{two people gather around a life size } \\ \textbf{chess game to {\color{red} celebrate life} .}} & \specialcell[t]{\textbf{three brown and black dogs are} \\\textbf{ splashing in the water {\color{blue} looking for mermaids} .}}\\
\hline
\\
\hline
\textbf{Example \#4}& \textbf{POLITICAL ({\color{blue} Democrat} to {\color{red} Republican})} & \textbf{POLITICAL ({\color{red} Republican} to {\color{blue} Democrat})}\\
\hline
\hline
SRC & \specialcell[t]{thank you for your commitment to a \\strong {\color{blue} public education system} , senator !} & \specialcell[t]{i absolutely agree with  \\{\color{red} senator paul's} actions .}\\
\hline
\hline
BT & \specialcell[t]{thanks for your vote for a \\ balanced budget amendment , sir !} & \specialcell[t]{i ' m merchandising with the\\ rhetoric of {\color{blue} senator warren} .}\\
\hline
\textbf{G-GST} &  \specialcell[t]{\textbf{thank you for your commitment to a }\\ \textbf{ strong {\color{red} conservative system} , governor !}} & \specialcell[t]{\textbf{i absolutely agree with {\color{blue} brian ' s} } \\\textbf{dire actions . .}}\\
\hline
\textbf{B-GST} &  \specialcell[t]{\textbf{thank you for your commitment to a strong }\\ \textbf{ {\color{red} constitutional system , senator scott} !}} & \specialcell[t]{\textbf{i absolutely agree with } \\\textbf{ {\color{blue} elizabeth warren ' s} actions .}}\\
\hline
\\
\hline
\textbf{Example \#5}& \textbf{GENDER ({\color{blue} Male} to {\color{red} Female})} & \textbf{GENDER ({\color{red} Female} to {\color{blue} Male})}\\
\hline
\hline
SRC & \specialcell[t]{this is a spot that ' s making very \\ {\color{blue}solid food} , with good quality product .} & \specialcell[t]{this a great place for {\color{red} a special date} or \\ to {\color{red}take} someone from out of town .} \\
\hline
\hline
\textbf{B-GST} & \specialcell[t]{\textbf{this is a cute spot that ' s making {\color{red}me very}} \\ \textbf{{\color{red}happy} , with good quality product .}} & \specialcell[t]{\textbf{this a great place for a {\color{blue}bachelor}} \\ \textbf{or to {\color{blue}meet} someone from out of town .}} \\
\hline

\end{tabular}
\end{center}
\caption{Examples of generated sentences to be compared down a column (B-GST and G-GST are our models, SRC is the input sentence). Attributes are colored. }
\label{sents-examples}
\end{table*}

\subsection{Result Analysis}
From human evaluations in Tables \ref{YAI-human-DR-BGST} and \ref{PG-human-DR-BGST}, we see that our models (specifically, \textbf{B-GST}) outperform state-of-art systems by a good margin on almost all parameters as judged by humans, across all datasets. More importantly, as Table \ref{sents-examples} shows, our models generate realistic and natural-sounding sentences while retaining core content - an aspect on which previous models seem to be seriously lacking. While our \textbf{G-GST} model does worse than \textbf{B-GST} due to a weak Retrieve mechanism, \textbf{G-GST} provides us a way to guide the generation and control attributes, making it more suitable for real-world applications after improving Retrieve in future.
We find that metrics based on learned models - perplexity and accuracy, do not correlate entirely well with human evaluations, an observation also shared by \citet{li-etal-2018-delete}. They are also heavily dependant on the distribution of data that they are trained on. A system that simply chooses a random sentence from the target training corpus as its output will score highly on both these metrics. For instance, the \textbf{BT} model in Table \ref{PG-auto-DR-BGST} has a high style but a considerably lower BLEU score than \textbf{B-GST}. It is important therefore, to not consider them in isolation. Further, human reference sentences themselves score poorly using both these metrics as shown in these tables. Manual inspection of classifier accuracies shows that these classifiers give unreliable outputs that do not match human ratings. This is the case with the \textbf{CA} model in Table \ref{YAI-human-DR-BGST}. Similar problems exist with regarding BLEU in isolation. A system that simply copies the source sentence will obtain high BLEU scores. 

GLEU, however seems to strike a balance between target style match and content retention, as it takes the source, reference as well as predicted sentence into account. We see that GLEU scores also correlate with our own human evaluations as well as those of \citet{li-etal-2018-delete}. While a detailed statistical correlation study is left for future work, the fact remains that GLEU is not susceptible to the weaknesses of other automatic metrics described above. Our uniformly state-of-art GLEU scores possibly indicate that we make only necessary edits to the source sentence. 

\begin{figure}[htp]
\centering
\includegraphics[width=7.5cm]{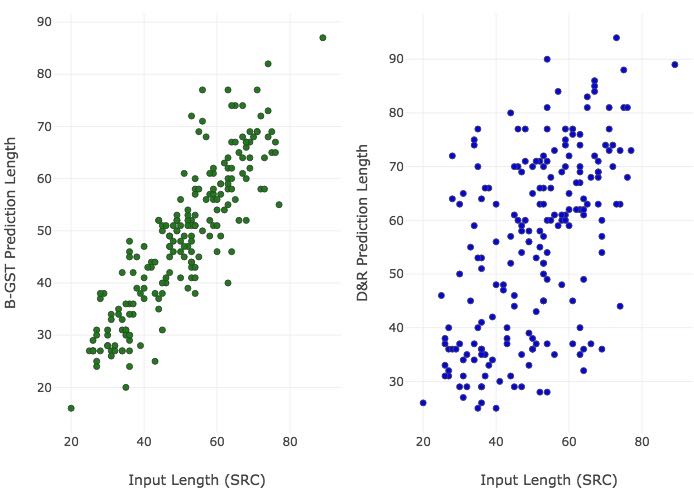}
\caption{Correlation of B-GST (ours, left) with input sentence lengths vs D\&R's (right) sentence lengths with input sentence lengths.}
\label{scatter}
\end{figure}

Keeping all the above considerations in mind, automatic metrics are still indicative and useful as they can be scaled to evaluate larger sets of models and datasets. From Tables \ref{YAI-auto-DR-BGST} and \ref{PG-auto-DR-BGST}, we see that we consistently outperform current state-of-art systems on BLEU. As shown by our high BLEU scores, one can conclude to some extent that our models retain non-stylistic parts well. Figure \ref{scatter} shows that unlike the current state-of-art \textbf{D\&R} model, the lengths of our generated sentences closely correlate with source sentence lengths. \textbf{B-GST} scores well on perplexity across datasets, a consistency that is not exhibited by any other model.

\section{Related Work}
One category of previous approaches is based on training adversarial networks to learn a latent representation of content and style. \citet{shen2017style} train a cross-aligned auto-encoder, with a shared content and separate style distribution. \citet{pmlr-v70-hu17e} use VAEs with attribute discriminators to learn similar latent representations. This approach has been later encapsulated in encoder-decoder frameworks \citep{fu2018style,John2018DisentangledRL,zhang-etal-2018-shaped, Zhang2018StyleTA}. Problems with these approaches have been discussed in the introduction.

Approaches that do not rely on a latent representation to separate content and attribute exist too. These include reinforcement learning based approaches \citep{xu-etal-2018-unpaired, gong2019reinforcement}, an unsupervised machine translation based approach \citep{s2018multipleattribute} and the DRG approach \citep{li-etal-2018-delete}. The former two approaches suffer from sparsity and convergence issues and hence generate sentences of low-quality.

Previous approaches to use attention weights to extract attribute significance exist \citep{feng-etal-2018-pathologies, Li2016UnderstandingNN, Globerson:2006:NTT:1143844.1143889}, including the salience deletion method of \citet{li-etal-2018-delete} but they do not perform well on understanding sentence context while choosing attributes, and do not leverage the contextual capacity of a Transformer. Lastly, \citet{dai2019style} describe the use of Transformers for style transfer in an adversarial generator-discriminator setting, by adding an additional style embedding to the transformer. We are unable to do a comparitive study as they do not yet publish their code or outputs. The same is the case for \citet{s2018multipleattribute}.

\section{Conclusion}
We propose the Generative Style Transformer that outperforms state-of-art systems on sentiment, gender and political slant. Our model leverages the DRG framework, massively pre-trained language models and the Transformer network itself. 

\section*{Acknowledgments}
The authors of this paper would like to thank Swati Tiwari, Nishant Thakur and Akshit Mittal for their help with evaluation of results, and the anonymous reviewers for their suggestions and comments.

\bibliography{emnlp-ijcnlp-2019}
\bibliographystyle{acl_natbib}

\end{document}